\documentclass{article}

\usepackage{arxiv}

\usepackage[utf8]{inputenc}
\usepackage{hyperref}       
\usepackage{url} 
\usepackage{amsfonts}
\usepackage{algorithm2e}
\usepackage{subcaption}
\usepackage{booktabs}
\usepackage{lmodern}
\usepackage{microtype}      
\usepackage{nicefrac}       
\usepackage{graphicx}
\usepackage{natbib}
\usepackage{doi}

\usepackage[flushleft]{threeparttable}  
\usepackage{siunitx}
\usepackage{appendix}
\newcommand{\fscore}{\text{F}_1\text{-score}}

\usepackage[dvipsnames]{xcolor}  
\definecolor{ballblue}{rgb}{0.13, 0.67, 0.8}
\definecolor{sapgreen}{rgb}{0.31, 0.49, 0.16}
\definecolor{candypink}{rgb}{0.89, 0.44, 0.48}

\title{\textsc{Tuna-AI}: tuna biomass estimation with Machine Learning models trained on oceanography and echosounder FAD data}
\date{}

\renewcommand{\shorttitle}{Tuna-AI: tuna biomass estimation with Machine Learning models}

\author{
    Daniel Precioso\\
    Department of Computer Science\\
    Higher School of Engineering\\
    Universidad de C\'adiz, Spain\\
    \texttt{daniel.precioso@uca.es}\\
    \And
    Manuel Navarro-García\\
    Universidad Carlos III de Madrid, Spain\\
    Komorebi AI Technologies,  Madrid, Spain\\
    \texttt{mannavar@est-econ.uc3m.es}\\
    \And
    Kathryn Gavira-O'Neill\\
    Satlink\\
    Madrid, Spain\\
    \texttt{kgo@satlink.es}\\
    \And
    Alberto Torres-Barrán\\
    Komorebi AI Technologies\\
    Madrid, Spain\\
    \texttt{alberto.torres@komorebi.ai}\\
    \And
    David Gordo\\
    Komorebi AI Technologies\\
    Madrid, Spain\\
    \texttt{david.gordo@komorebi.ai}
    \And
    V\'ictor Gallego-Alcal\'a\\
    Komorebi AI Technologies\\
    Madrid, Spain\\
    \texttt{victor.gallego@komorebi.ai}
    \And
    David G\'omez-Ullate\thanks{Corresponding author. On leave of absence from Department of Theoretical Physics, Universidad Complutense de Madrid, Spain.}\\
    Department of Computer Science\\
    Higher School of Engineering\\
    Universidad de C\'adiz, Spain\\
    \texttt{david.gomezullate@uca.es} \\
}

\begin{document}

\maketitle

\begin{abstract}
Echo-sounder data registered by buoys attached to drifting FADs provide a very valuable source of information on populations of tuna and their behaviour. This value increases when these data are supplemented with oceanographic data coming from CMEMS. We use these sources to develop \textsc{Tuna-AI}, a Machine Learning model aimed at predicting tuna biomass under a given buoy, which uses a 3-day window of echo-sounder data to capture the daily spatio-temporal patterns characteristic of tuna schools. As the supervised signal for training, we employ more than \num{5000} set events with their corresponding tuna catch reported by the AGAC tuna purse seine fleet. 
\end{abstract}


\section{Introduction}

Throughout tropical and sub-tropical oceans, a variety of fish species are known to aggregate around objects drifting on the surface, a behavior which fishermen have learned to exploit for centuries~\citep{Castro2002AHypothesis,Maufroy2015Large-ScaleOceans}. In tropical tuna purse-seine fisheries, targeting mainly skipjack tuna (\textit{Katsuwonus pelamis}), yellowfin tuna (\textit{Thunnus albacares}) and bigeye tuna (\textit{Thunnus obesus}), these drifting objects, known as drifting Fish Aggregating Devices (dFADs), have become an essential tool for locating tuna-schools and increasing fishing efficiency. Today, more than $55\%$ of tropical tuna caught by industrial purse-seine vessels in the Indian, Atlantic and Pacific oceans is caught using dFADs, accounting for $36\%$ of the world’s total tropical tuna catch \citep{Wain2021QuantifyingFisheries,ISSF2021Status2021}.

Initially, dFADs were of natural origin, such as floating logs or objects, that fishermen would come across while searching for free-swimming schools of tuna. In the mid-1980s, tools began to be developed to allow for tracking of these dFADs, and fishermen themselves designed purpose-built dFADs that could be attached to tracking beacons: first based on radar reflectors or radio, and later satellite connected GPS buoys, allowing the dFADs to be located remotely~\citep{Davies2014TheOcean,Lopez2014EvolutionOceans}. The use of these tracking buoys has been considered “the most significant technological development that has occurred (…) for increasing the efficiency of dFAD tuna fishing”~\citep{Lopez2014TheFisheries}. Nowadays, most dFADs are equipped with satellite-linked instrumented buoys which include both GPS and an echo-sounder, providing fishermen with accurate geolocation information as well as an estimate of associated tuna biomass. These buoys allow fishing crews to monitor remotely their dFADs and the biomass they aggregate in real-time, so they can target those with larger aggregated schools, thus increasing their catch while reducing searching effort~\citep{Lopez2014EvolutionOceans,Molina2003Statistics1984-2002}.

The widespread use of dFADs has led to large-scale changes in industrial purse-seine fishing fleets targeting tropical tunas, affecting traditionally used indices of Catch Per Unit Effort (CPUE) such as search-time and time-at-sea~\citep{Fonteneau2000AFADs}. In this context, some authors have highlighted the need for fishery-independent abundance indices and the use of non-traditional data sources to monitor tuna stock health and the effects of fishing pressure over time~\citep{Baidai2020MachineData,Santiago2016TowardsTT-BAI,Santiago2020ABuoys}. The echo-sounder buoys attached to dFADs across the world’s oceans can be set to transmit frequent geo-referenced biomass estimates. Given the number and wide distribution of dFADs in recent years, the information provided by these echo-sounder buoys could be very valuable. ~\citet{Santiago2016TowardsTT-BAI} presented the first Buoy-Derived Abundance Index (BAI) for tropical tunas as a proxy of CPUE, based on the biomass estimates provided by three echo-sounder buoy brands in the Atlantic, Indian and Pacific Oceans. 

 However, several authors have reported substantial differences between the biomass estimates provided by echo-sounder buoys and observed biomass~\citep{Lopez2016ADevices,Escalle2019ReportData,Orue2019FromBuoys}, evidencing the variable nature of fish aggregations under dFADs, often made up of pelagic species other than tuna~\citep{Castro2002AHypothesis}. Likewise, the influence of oceanic conditions on fish distribution and behavior likely drives aggregation patterns of tuna around dFADs~\citep{Lopez2017EnvironmentalBuoys,Druon2017SkipjackOceans,Schaefer2007MovementsData}. Therefore, in order to develop a representative index of abundance from echo-sounder buoy data, it is also important to consider and understand the effect of these variables on the biomass estimates given by these echo-sounder buoys.

Although some studies have already compared biomass estimates from the buoys to catch data~\citep{Baidai2020MachineData,Lopez2016ADevices,Mannocci2021MachineFisheries}, the approach in this paper is the first to consider oceanographic data as predictor variables in Machine Learning models. Likewise, others have combined oceanographic variables and catch data, without using echo-sounder buoy information~\citep{Druon2017SkipjackOceans}. Lastly, others have considered the effects of oceanographic conditions on buoy biomass estimates, without directly comparing it to catch data~\citep{Lopez2017EnvironmentalBuoys, Santiago2020ABuoys}. In this sense, the current study aims to evaluate several models and methods in order to find which can most accurately estimate biomass under echo-sounder buoys, combining information from all three sources: catch data, oceanographic variables, and echo-sounder buoy information, including positional data and biomass estimates. 


\section{Material and methods}

\subsection{Database description}

Our study draws from three sources of information: activity data on FADs, echo-sounder buoy data, and oceanography data.

\subsubsection{FAD logbook data} \label{subsubsec: FAD logbook data}

The first source of information corresponds to the registered interactions between fishing vessels and echo-sounder buoys, obtained from the FAD logbooks of the Spanish tropical tuna purse seine fleet operating in the Atlantic, Indian and Pacific Oceans (2018 - 2020, AGAC\footnote{Asociaci\'on de Grandes Atuneros Congeladores} ship owner's association data). This FAD logbook dataset contains almost \num{66000} interactions with Satlink buoys. Each record within the dataset can be traced to a specific buoy (using the ID and model of the buoy attached to each dFAD), and  contains information about the date, time and GPS coordinates where the interaction occurred, as well as the nature of the interaction (see~\citet{Ramos2017SPANISHREQUIREMENTS} for definitions and descriptions of each interaction). 

For the purposes of the current study, only events registered as "Set" and "Deployment" were used  ~\citep{Ramos2017SPANISHREQUIREMENTS}. Set events, and their associated catch data, were used as ``positive cases" and considered an accurate representation of real tuna biomass under a given dFAD. This follows the assumption that the entire fish aggregation present at the dFAD is captured by the vessel during the set. This assumption can be validated by studying the echo-sounder signal before and after the set event. Deployment events were used as ``negative cases", whereby we consider that no tuna is present under newly deployed dFADs~\citep{Orue2019UsingOcean}.

\subsubsection{Echo-sounder buoy data}   \label{subsubsec: Echo-sounder buoy data}
\begin{figure}[htbp]
    \centering
    \includegraphics[width=.6\linewidth]{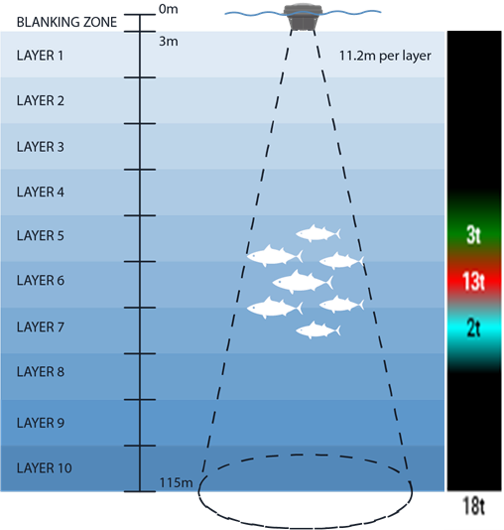}
    \caption{Left: Depth layer configuration and set-up of the Satlink echo-sounder buoys. Right: example of the biomass estimates (in metric tons) and echogram display available to buoy users. Raw acoustic backscatter is converted into biomass estimates based on the target strength of skipjack tuna \textit{(Katsuwonus pelamis)} using manufacturer's algorithms.}
    \label{fig:buoy_echo-sound}
\end{figure}

The echo-sounder buoy data was collected from \num{15497} Satlink buoys\footnote{SATLINK, Madrid, Spain, www.satlink.es} for which there were registered interactions in the FAD logbook data (see~\ref{subsubsec: FAD logbook data}). This database contained over 68 million records corresponding to buoys attached to dFADs scattered over the Atlantic, Indian and Pacific Oceans. Each record is referenced to a specific buoy ID and timestamp, ranging from 2018 to 2020, and contains biomass estimates (explained below) and GPS coordinates of the buoy's last known position at the time of measuring. Three buoy models (ISL+, SLX+ and ISD+) were used in this study, the details of which are described in Table~\ref{tab:buoy}. For all buoys, observation range of the echo-sounder is from 3m to 115m depth, split into ten layers, each with a resolution of 11.2m (see Figure~\ref{fig:buoy_echo-sound}). Biomass estimates (in metric tons, \si{\tonne}) are obtained from acoustic samples taken periodically throughout the day (see Table~\ref{tab:buoy}), and the average back-scattered acoustic response is converted into estimated tonnage, based on the target strength of Skipjack tuna (\textit{Katsuwonus pelamis}). To reduce the amount of information to be transmitted, only one measurement per hour is selected, corresponding to that with the highest estimated tonnage, and sent to the vessel at specific intervals during the day. Thus, the final temporal resolution of echo-sounder records for each buoy is 1h. The buoys have an internal detection threshold of 1 ton, which means that a total biomass estimation for any given measurement below 1 ton is not transmitted and thus interpreted as a zero-reading. 

\begin{table}[htbp]
 \centering
\begin{threeparttable}
 \begin{tabular}{l l p{.1\linewidth} p{.14\linewidth} p{.3\linewidth}}
 \toprule
\textbf{Model} & \textbf{Echo-sounder} & \textbf{Freq. (kHz)} & \textbf{Beam \newline Angle ($^\circ$)} & \textbf{Measuring rate} \\
 \midrule
 ISL+ & ES12 & 190.5 & 20 & Every 15 minutes \\
 SLx+ & ES16 & 200 & 23 & From sunrise to sunset:\newline every 5 minutes\newline From sunset to sunrise: every 60 minutes \\
 ISD+ & ES16x2 & 200 and 38\tnote{a} & 23 and 33 & From sunrise to sunset: \newline every 5 minutes \newline From sunset to sunrise: \newline every 60 minutes \\
 \bottomrule
\end{tabular}
\begin{tablenotes}
\item[a] \footnotesize{Biomass estimates are calculated according to the acoustic response registered by the 200kHz echo-sounder, allowing data to be comparable across all buoy models.}
\end{tablenotes}
\end{threeparttable}
\caption{Buoy models and characteristics}
\label{tab:buoy}
\end{table}

\subsubsection{Oceanography data} \label{subsubsec: oceanography}
Oceanographic data was downloaded from the global ocean model (products GLOBAL-ANALYSIS-FORECAST-PHY-001-024, $1/12^\circ$ resolution; and GLOBAL-ANALYSIS-FORECAST-BIO-001-028, $1/4^\circ$ resolution) provided by the EU Copernicus Marine Environment Monitoring Service\footnote{http://marine.copernicus.eu/}~\citep{GlobalMonitoringandForecastingCenter2018OperationalInformation}. For each record contained in the echo-sounder buoy data (see~\ref{subsubsec: Echo-sounder buoy data}), the following variables were downloaded: temperature (in $\rm ^\circ C$), chlorophyll-a concentration (in $\rm mg / m^3$), dissolved oxygen concentration (in $\rm mmol / m^3$), salinity (in psu), thermocline depth (calculated as the depth where water temperature is $2^\circ$C lower than surface temperature, in m), current velocity (in $\rm m/s$) and sea surface height anomaly (SSHa, deviation of the sea surface height from long term mean, in $\rm m$). All variables, except thermocline and SSHa, were downloaded at surface level (depth = $0.494$ m).

\subsection{Data preprocessing}

\subsubsection{Data merging} \label{subsubsec: data merging}
Each set and deployment event registered in the FAD logbook data has been linked to a specific buoy, using the buoy model and ID. Therefore, each record contained in the echo-sounder buoy database could be related to a given event. Furthermore, each record contained in the echo-sounder buoy database included the GPS coordinates of the buoy's last known position (LKP). Oceanographic information was thus collected for the date and position of each echo-sounder buoy record. Since oceanographic data are available on a grid with $0.08^\circ$ or $0.25^\circ$ resolution, we incorporate the data from the closest point on the grid to the buoy's position. Oceanographic variables change on a larger spatial scale compared to the grid spacing and buoys hourly movement, so no significant errors are incurred in this approximation.

\subsubsection{Echo-sounder window} \label{subsubsec: echo-sounder window}
\begin{figure}[htbp]
    \centering
    \includegraphics[trim={3cm 0cm 3cm 0cm},clip, width=\linewidth]{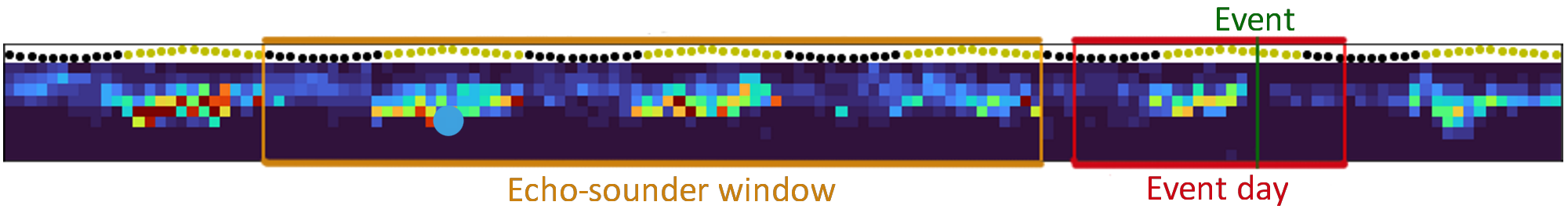}
    \caption{Example of how the 72h echo-sounder window (yellow box) is selected with respect to the event day  (red box). The set event registered in the FAD logbook (assuming UTC) is depicted as a green line. Sun inclination and day and night patterns are represented above the graph (yellow circles: day; black circles: night). Colored squares represent the biomass estimates for each measurement from the echo-sounder buoy. Spatio-temporal patterns of tuna activity are clearly visible in this plot.}
    \label{fig:echo-sound}
\end{figure}

It is known that tuna schools have well defined circadian behaviours around dFADs that we expect to see reflected in spatio-temporal patters in the echo-sounder signal. Typically, tuna arrive at dFADs at or near dawn, and depart around sunset, remaining near the dFAD for several days in a row~\citep{Forget2015BehaviourTelemetry, Dagorn2007BehaviorFADs}. To capture this regular behaviour, and in order to accurately compare data across time zones, all echo-sounder and set data was referred to solar time, for which we calculated the sun's inclination for each given position and time within the dataset. This also allowed us to calculate the time of sunrise and sunset per day in the echo-sounder window. There was considerable uncertainty around the time-zone of registered set and deployment events considered in the FAD logbooks. For this  reason, we have chosen to ignore the time of the event, and keep only the registered event date (see Figure~\ref{fig:echo-sound}).

To relate each set and deployment event to the data included in the echo-sounder buoy dataset, we retrieved all echo-sounder buoy data on a given window before a set and after a deployment event. Given the periodic behaviour of tuna mentioned previously, we tested the effect of including different length windows of echo-sounder data on the machine learning models (see ~\ref{subsubsec: ML algorithms}) included in our analyses: 24h, 48h and 72h. The echo-sounder window length with the best results would then be used for all following analyses. For set events, the last echo-sounder record included in the window corresponds to the sunset on the day prior to the event. For deployment events, the last echo-sounder record corresponds to 24h, 48h or 72h after sunset on the day of the last record included in the deployment window (see Figure~\ref{fig:echo-sound} for an example).

\subsubsection{Data cleaning}

Data used in this study can be quite noisy and it often contains errors, specially event data coming from the FAD logbook, since it is recorded by persons that tend to approximate the different measurements. For that reason it is crucial to clean the data to get rid of as many inconsistencies as possible. For a set or deployment event to be included in the final dataset, the following conditions had to be met:
\begin{itemize}
    \item The buoy ID registered in the FAD logbook data must match the buoy ID in the echosounder database, i.e. we ensure that echosounder data are available for the FAD on which the event took place.
    
    \item The echo-sounder window and event day for each buoy (see Figure~\ref{fig:echo-sound}) should not overlap. The reason to impose this condition is to ensure that inside the window used for prediction there has been no human intervention on the FAD.
    \item Following the same procedure as \citep{Escalle2019ReportData}, events with invalid positions (i.e. buoys on land) were removed from the dataset. 
    \item Events or measurements registered at positions with less than 200m water depth were discarded. This avoids including measurements that are potentially influenced by the sea-bed. 
  \item Using the last known position of the buoys, we computed buoy speed for each position, and dropped events and measurements where buoy speed was higher than 3 knots, since the surface currents in the tropical oceans rarely exceed this speed~\citep{Orue2019FromBuoys}. This avoids including measurements taken on-board a vessel and not representative of a dFAD.
\end{itemize}

After merging and filtering, the final dataset contained over \num{12000} events (see Table~\ref{tab:activity_eco}). These occurred on \num{10063} buoys, for which over \num{665000} records were collected.

\begin{table}[htbp]
\centering
\begin{tabular}{lS[table-format=5]S[table-format=5]S[table-format=5]S[table-format=5]}
\toprule
& {Atlantic} & {Indian} & {Pacific} & {Total} \\ 
\midrule
Set & 1500 & 2727 & 974 & 5201 \\
Deployment & 1369 & 2199 & 3426 & 6994 \\
\midrule
Total & 2869 & 4926 & 4400 & 12195 \\ 
\bottomrule \\
\end{tabular}
\caption{Number of events remaining after merging echo-sounder and FAD logbook data.}
\label{tab:activity_eco}
\end{table}

\subsection{Model testing}

We tested several models  using varying feature sets, in order to assess the relative contribution of different features to model accuracy, as well as test the overall performance of different modelling methods.

\subsubsection{Feature selection}

\begin{figure}[htbp]
    \centering
    \includegraphics[width=.8\linewidth]{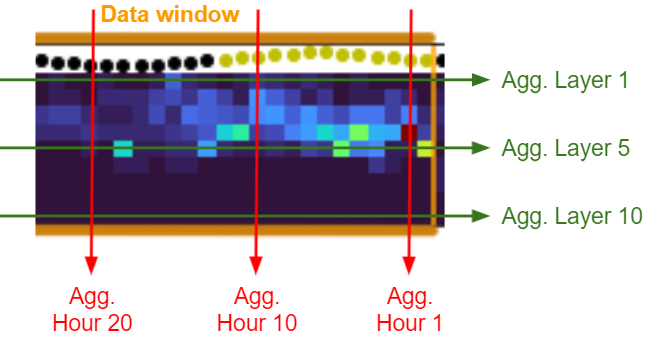}
    \caption{Visual example of how the biomass measurements are aggregated.}
    \label{fig:echo_agg}
\end{figure}

Based on the merged dataset (see~\ref{subsubsec: data merging}), we considered the variables included in Table~\ref{tab:features} as features to be included in each model. The original biomass measurements form a $10 \times \{24, 48, 72\}$ matrix, depending on the echo-sounder window size. These values are not fed directly into the models; instead they are aggregated both by row (layers) and column (hours) using the maximum, as shown in Figure~\ref{fig:echo_agg}. These aggregations are then used directly as features for the different models.

As for the number of zero-readings, we just count how many missing records there are in the echo-sounder window. These missing values are imputed to 0, since the echo-sounder does not transmit any data when the measurement is below 1\si{\tonne}.

\begin{table}[htbp]
\centering
\begin{tabular}{lccc}
\toprule
 & Echo & Echo + Ocean & All \\
 \midrule
{Biomass measurements} & \checkmark & \checkmark & \checkmark \\
{Number of zero-readings} & \checkmark & \checkmark & \checkmark \\
{Buoy model} & \checkmark & \checkmark & \checkmark \\
\midrule
{Chlorophyll-a} &  & \checkmark & \checkmark \\
{Dissolved oxygen} &  & \checkmark & \checkmark \\
{Salinity} &  & \checkmark & \checkmark \\
{Thermocline depth} &  & \checkmark & \checkmark \\
{Temperature} &  & \checkmark & \checkmark \\
{Current velocity} &  & \checkmark & \checkmark \\
{SSHa} &  & \checkmark & \checkmark \\
\midrule
{Day and month} &  &  & \checkmark \\
{Year} &  &  & \checkmark \\
{Latitude} &  &  & \checkmark \\
{Longitude} &  &  & \checkmark \\
{Ocean basin} &  &  & \checkmark \\
{Sunrise hour} &  &  & \checkmark \\
{Sunset hour} &  &  & \checkmark \\
\bottomrule\\
\end{tabular}
\caption{Grouped features used for the models. ``Echo'' includes only data relating to echo-sounder measurements from the echo-sounder buoy database. ``Echo + Ocean'' includes oceanography data for the position and time of each record in the echo-sounder buoy database. ``All'' contains further derived data from the position and time of each record in the echo-sounder buoy database.}
\label{tab:features}
\end{table}

\subsubsection{Model selection}
Models were trained to achieve three different tasks:
\begin{enumerate}
    \item A binary classification task, where the target variable $y$ (tuna biomass) can assume the values \textit{absent} ($y<10 \si{\tonne}$) or \textit{present} ($y \geq 10 \si{\tonne}$).
    \item A ternary classification task,  where the target variable  (tuna  biomass) can assume the values \textit{low} ($y<10 \si{\tonne}$), \textit{medium} ($10 \si{\tonne} \leq y<30 \si{\tonne}$) or \textit{high} ($y \geq 30 \si{\tonne}$).
    \item A regression task, where we directly estimate the tuna biomass $y$, in tons.
    \item A threshold regression task, where we directly estimate the tuna biomass $y$, in tons, up to a threshold of $100 \si{\tonne}$. Estimations equal or higher than that were grouped together as $\geq 100 \si{\tonne}$.
\end{enumerate}

The thresholds to define the categories were chosen according to various criteria. In both classification tasks, the lower threshold was based on best-practice guidelines to decrease shark bycatch, which recommend avoiding sets on tuna schools less than $10$ tons~\citep{Restrepo2016MitigationFisheries}. In the ternary classification task, the class limit for \textit{medium} was established according to the median catch ($30$ tons) in the dataset. In the threshold regression task, we selected 100 tons since sets above that are relatively rare (\num{315} events, $8.1\%$).

\subsubsection{Machine Learning algorithms and training}  \label{subsubsec: ML algorithms}
Following the usual approach in supervised Machine Learning, we split the dataset into training ($75\%$, \num{9152} events, \num{3893} sets and \num{5259} deployments) and test (25\%, \num{3051} events, \num{1309} sets and \num{1742} deployments) preserving  the total class distribution. We considered the performance of a baseline rule-based model and four different ML models in the classification and regression tasks:

\begin{itemize}
    \item Baseline: The baseline model only uses the biomass estimates from records contained in the echo-sounder window (see~\ref{subsubsec: echo-sounder window}) for each event. It arrives at an overall biomass estimation by applying a set of aggregation rules on this matrix. The possible candidates for these aggregation rules are the sum, the mean and the maximum value. For each task, the selected baseline model was the one with the best performance.  
    \item Logistic Regression classifier (LR): a linear model for the classification task
    \item  Elastic Net regressor (ENet): for the regression task, with three regularization techniques, namely $L1$ penalization, $L2$ penalization and elastic net. 
    \item Random Forest (RF) algorithm~\citep{Breiman2001RandomForests}.
    \item Gradient Boosting (GB) algorithm~\citep{Friedman2001GreedyMachine}.
    \item XGBoost (XGB) algorithm~\citep{Chen:2016:XST:2939672.2939785}.
\end{itemize}

For training and evaluating the models, we used the corresponding algorithms implemented in the python \texttt{scikit-learn} \citep{Pedregosa2011Scikit-learn:Python} and \texttt{XGBoost} \citep{Chen:2016:XST:2939672.2939785} libraries. Each model was trained on three different sets of predictor variables, listed in Table~\ref{tab:features}. 

\subsubsection{Model evaluation}
For each model, we performed a grid search with 5-fold cross-validation to find the optimal hyper-parameters. To select the best set of hyper-parameters for each model, we use the Area Under the Receiver Operating Characteristic Curve (ROC AUC) for the classification tasks and the Mean Absolute Error (MAE) for the regression tasks. AUC is defined by plotting the ROC curve (graphing the true positive rate against the false negative rate at several thresholds) and computing the area below the curve. MAE score is defined as the average of the absolute values of the errors when comparing the observed and the predicted values. 

To report the performance of the models in the classification tasks, we choose the $\fscore$, which is the harmonic mean of precision and recall. For the multi-class task, we report the weighted averaged $\fscore$.


\section{Results}

\subsection{Variable echo-sounder window}

\begin{table}[htbp]
 \centering
 \begin{tabular}{lcccc}
 \toprule
       & \multicolumn{2}{c}{Classification ($\fscore$)} & \multicolumn{2}{c}{Regression (MAE)}\\
       \cmidrule(lr){2-3} \cmidrule(lr){4-5}
 Hours & Binary & Three class & Standard & Threshold \\
 \midrule
 24 & 0.911 & 0.811 & 10.16 & 8.70 \\
 48 & 0.919 & 0.813 & 10.05 & 8.63 \\
 72 & \textbf{0.925} & \textbf{0.824} & \textbf{10.03} & \textbf{8.54} \\
 \bottomrule \\
\end{tabular}
\caption{Model score according to echo-sounder window size for Gradient Boosting regression and classification models.}
\label{tab:hours}
\end{table}

All GB models, regardless of task, were improved with extended echo-sounder windows. Within the classification tasks, the best overall results were achieved by the binary classification model ($\fscore = 0.925$, Table~\ref{tab:hours}) using the 72h echo-sounder window. Similar results are shown for the regression tasks, where models using the 72h echo-sounder window had the lowest MAE (Table~\ref{tab:hours}). Of the two regression tasks, the threshold regression performed better, with MAE almost 1.5 tons lower than standard regression (Table~\ref{tab:hours}).

As echo-sounder windows spanning 72h showed the best result across all models, this is the echo-sounder window considered in the following analyses. 

\subsection{Classification tasks}

The best performing model in both classification tasks is GB. Performance also increases for every model as the number of features included in the training increases, i.e. when the models are able to learn from a larger set of features (Table~\ref{tab:classification}). Thus, the highest overall accuracy score was achieved by the binary task GB model trained with all features ($\fscore = 0.925$, Table~\ref{tab:classification}). The least accurate results were achieved by the ternary classification Baseline model, which was almost 20\% less accurate than the best performing model for this task, the GB model with all features.

\begin{table}[htbp]
\centering
\begin{tabular}{lcccccc}
\toprule
       & \multicolumn{3}{c}{Binary} & \multicolumn{3}{c}{Three class} \\
       \cmidrule(lr){2-4} \cmidrule(lr){5-7}
Models & Echo & Echo + Ocean & All & Echo & Echo + Ocean & All \\
\midrule
Baseline & 0.754 & - & - & 0.648 & - & -\\
LR  & 0.885 & 0.889 & 0.895 & 0.773 & 0.788 & 0.799\\
RF  & 0.893 & 0.911 & 0.918 & 0.794 & 0.799 & 0.807\\
XGB & 0.900 & 0.913 & 0.922 & 0.798 & 0.805 & 0.813\\
GB & 0.907 & 0.924 & \textbf{0.925} & 0.791 & 0.812 & \textbf{0.824}\\
\bottomrule \\
\end{tabular}
\caption{$\fscore$s for test events (classification task)}
\label{tab:classification}
\end{table}

\begin{figure}[htbp]
    \centering
    \includegraphics[width=.96\linewidth]{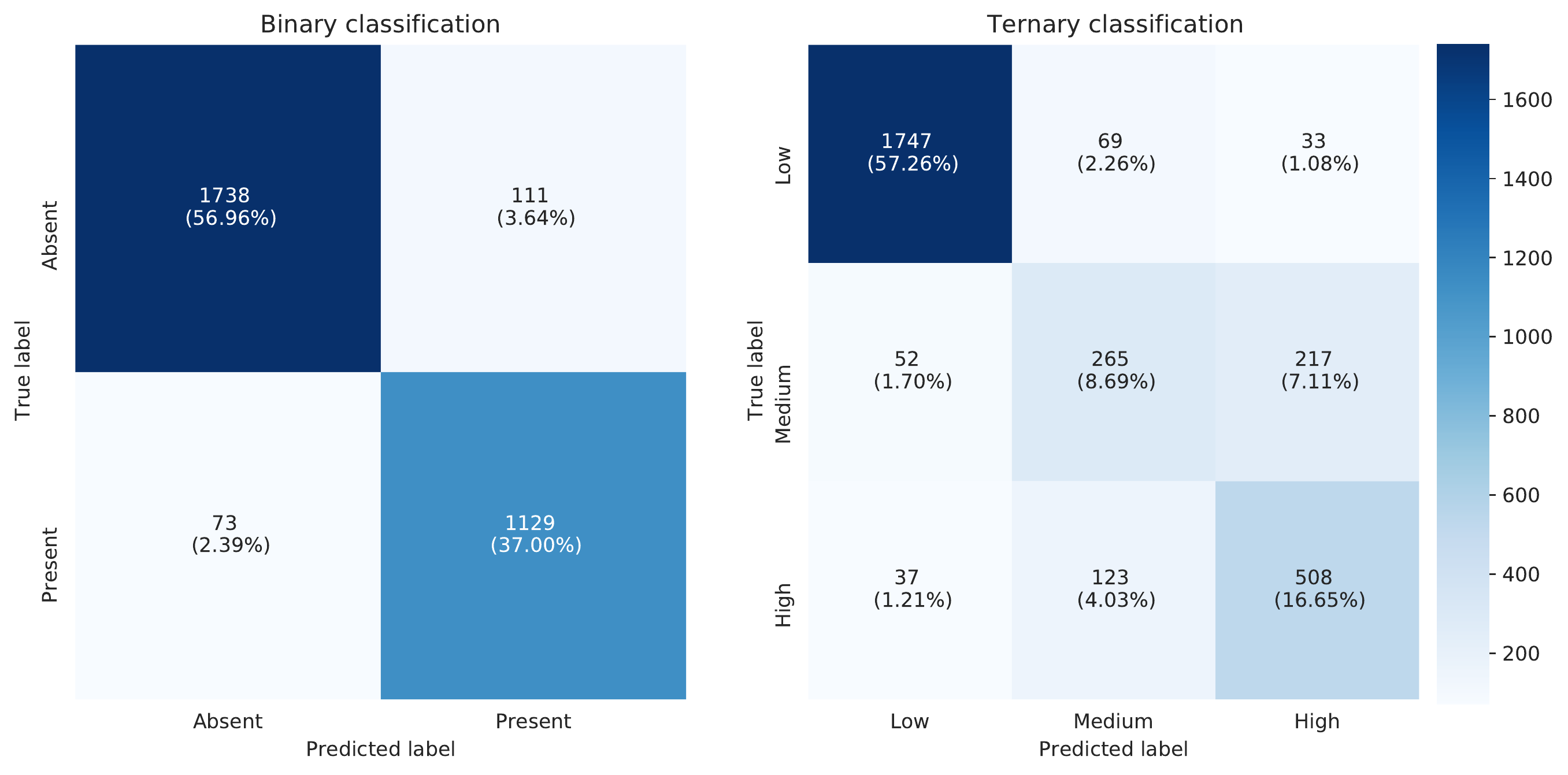}
    \caption{Confusion matrices with the performance of the best model on the test set.}
    \label{fig:cm}
\end{figure}

When analysing the confusion matrix for the test set of both classification tasks (Figure ~\ref{fig:cm}), we see that the GB model in the binary classification task has a high success rate in classifying whether tuna is present or absent, misclassifying results in only 6.03\% of cases (Figure~\ref{fig:cm}). However, the ternary classification GB model finds it harder to discriminate between medium ($10\si{\tonne} \leq y < 30 \si{\tonne}$) and high ($y \geq 30 \si{\tonne})$ biomass estimations, having misclassified results in these two classes in 11.14\% of cases (Figure~\ref{fig:cm}).

\subsection{Regression tasks}

Regarding the regression task, the results obtained by all the models trained on the different sets of predictor variables are shown in Table~\ref{tab:regression}. As in the classification tasks, the GB model for regression tasks also showed the overall best performance. More specifically, the threshold regression GB model was the most accurate, achieving a MAE nearly 3 tons lower than the baseline model for the same task, and 1.49 tons lower than the GB model for the standard regression (Table~\ref{tab:regression}). It is also noteworthy that, as for the classification tasks, all models benefited from the inclusion of position and oceanography data, and were able to use this information to improve their predictions with respect to models that were only fed the echo-sounder data.

\begin{table}[htbp]
\centering
\begin{tabular}{lcccccc}
\toprule
       & \multicolumn{3}{c}{Regression} & \multicolumn{3}{c}{Regression (Threshold)} \\
       \cmidrule(lr){2-4} \cmidrule(lr){5-7}
Models & Echo & Echo + Ocean & All & Echo & Echo + Ocean & All \\
\midrule
Baseline & 12.85 & - & - & 11.40 & - & -\\
ENet & 13.99 & 13.70 & 13.52 & 12.18 & 11.84 & 11.60\\
RF  & 10.74 & 10.30 & 10.20 & 9.42 & 8.93 & 8.84\\
XGB & 11.37 & 10.86	& 10.76 & 9.60 & 9.13 & 9.02\\
GB & 10.51 & 10.10 & \textbf{10.03} & 9.18 & 8.74 & \textbf{8.54} \\
\bottomrule \\
\end{tabular}
\caption{MAE scores for test events (regression task)}
\label{tab:regression}
\end{table}

The mean average error (MAE) shown in Table~\ref{tab:regression} hides an important fact: the errors are very  different for the two events included in the test set. Indeed, deployment events have by definition an observed biomass of zero: when tested over deployment events, the GB model has a MAE of $1.23$ tons, while the MAE for set events is $21.66$ tons. The reported overall MAE of $10.03$ tons is thus the weighted average of these different populations. This number is strongly dependent on how the training dataset has been designed (see Table~\ref{tab:activity_eco}).

\begin{figure}[htbp]
    \centering
    \includegraphics[width=.57\linewidth]{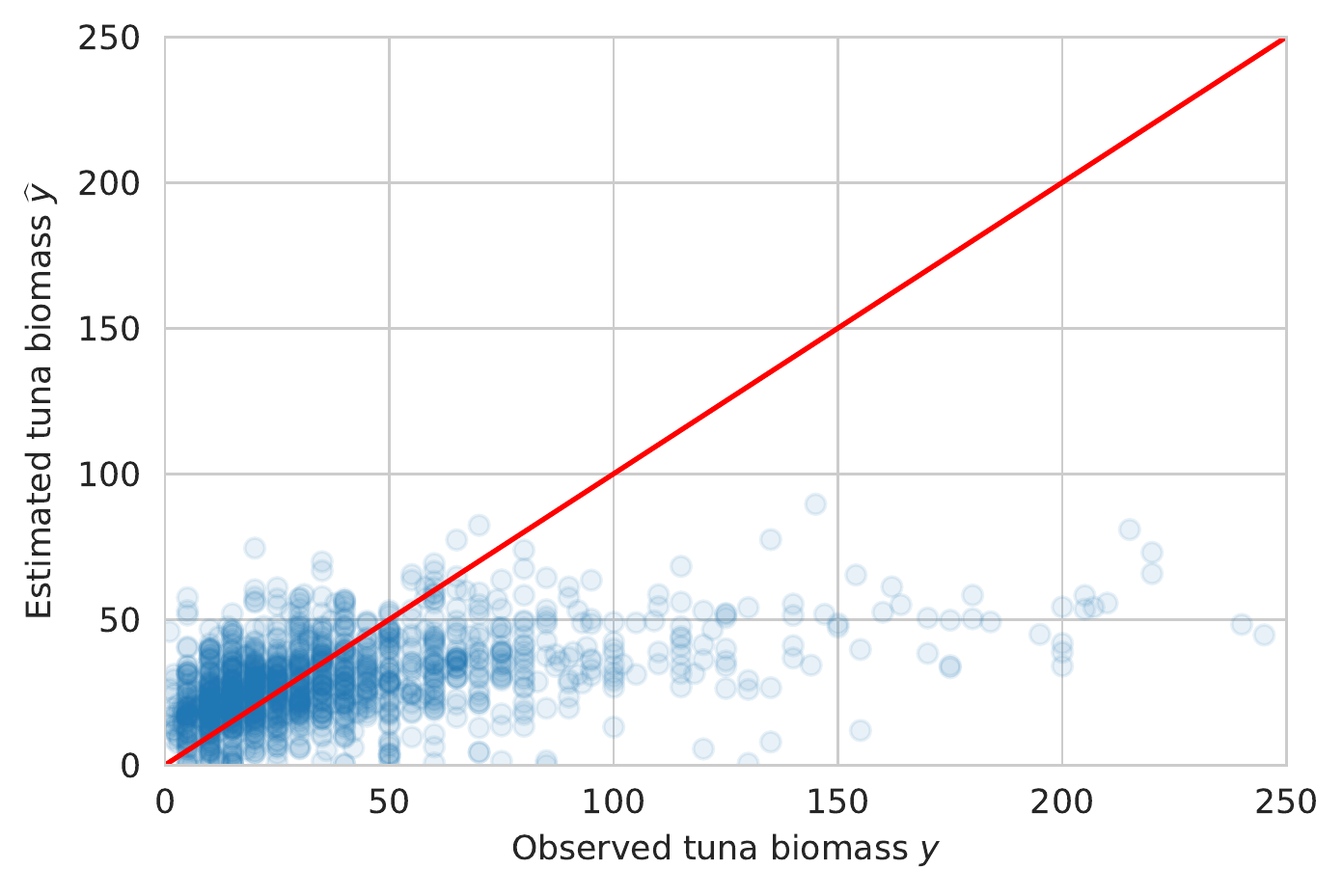}
    \hfill
    \includegraphics[width=.41\linewidth]{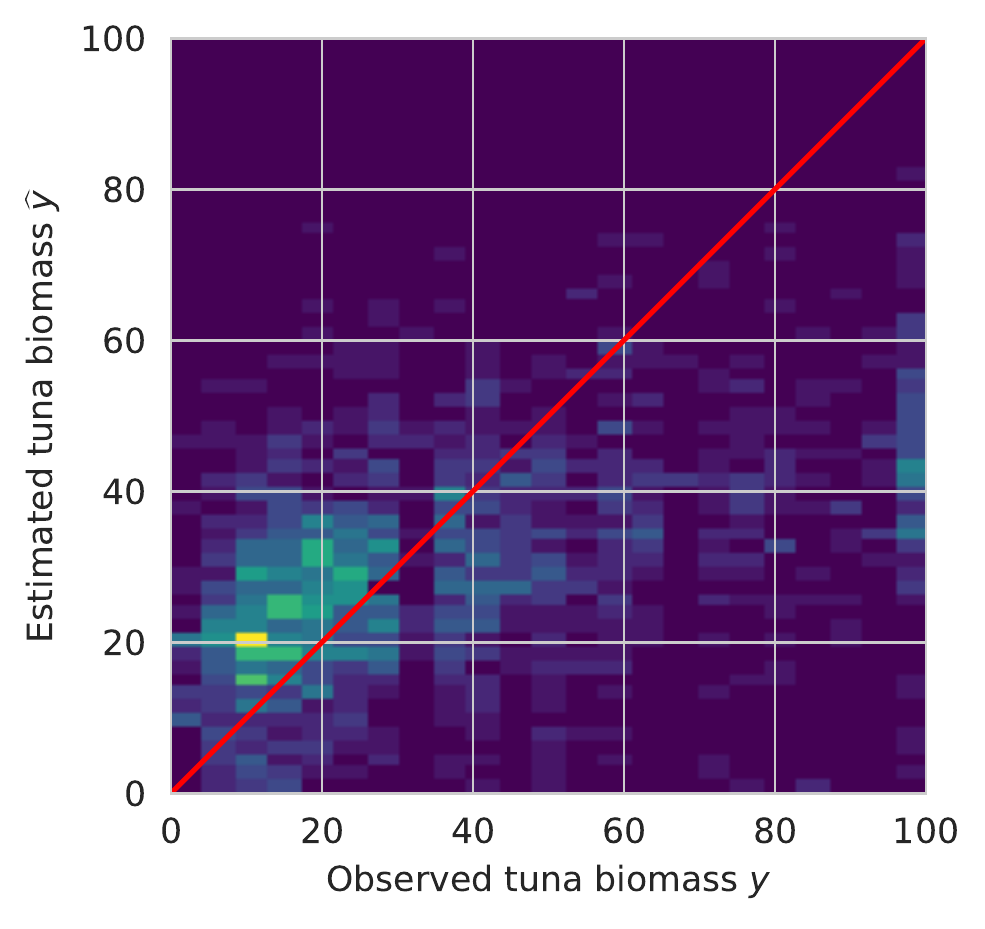}
    \caption{Observed tuna biomass against estimated biomass in set events. Scatter plot for the regression task (left) and density plot for the regression threshold task (right).}
    \label{fig:reg_errors}
\end{figure}

\begin{figure}[htbp]
    \centering
    \includegraphics[width=.65\linewidth]{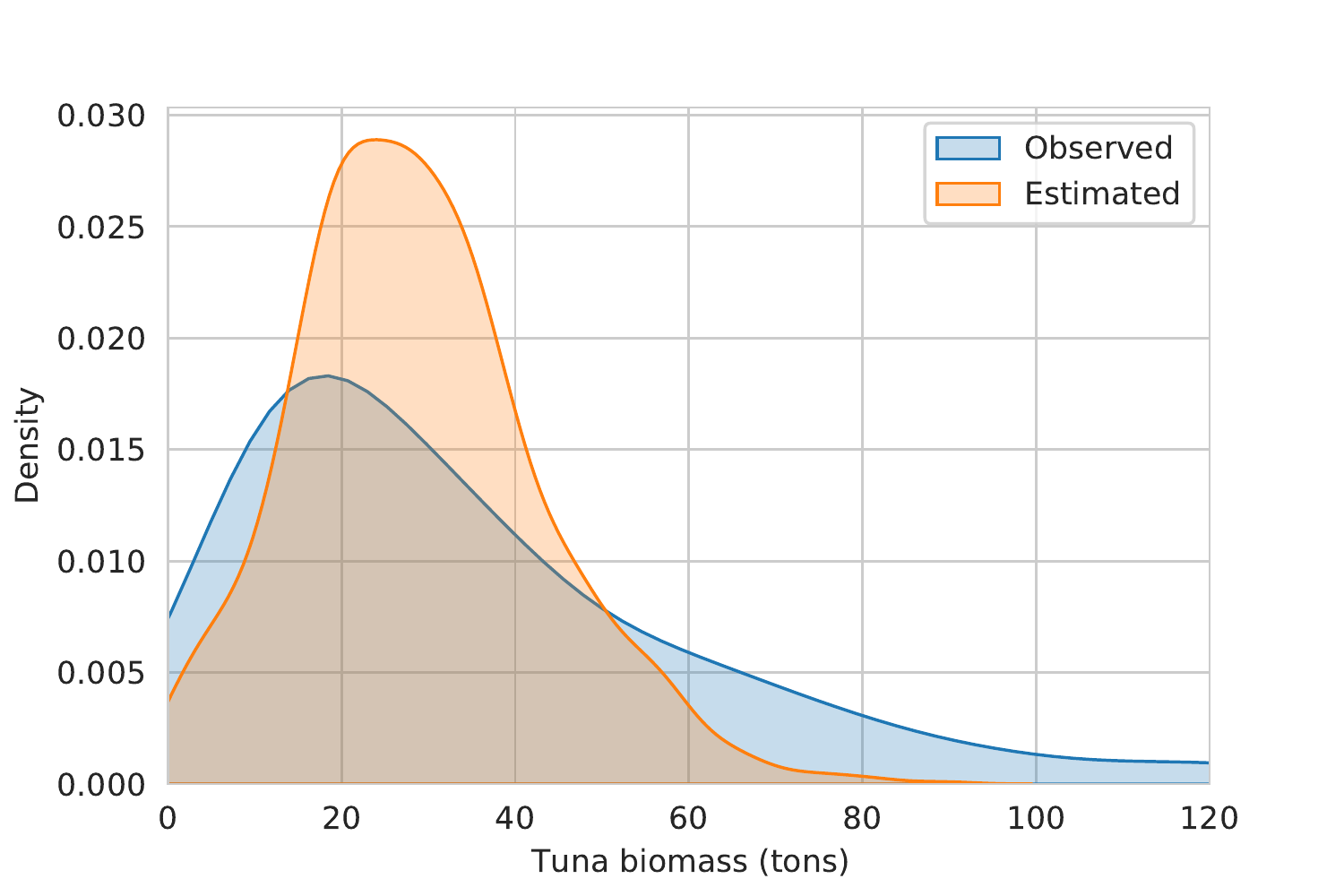}
    \caption{Density distributions (set events) of observed and estimated tuna biomass with Gradient Boosting model.}
    \label{fig:density_dist}
\end{figure}


When looking more closely at the predictions of the best regression model, shown in Figure~\ref{fig:reg_errors} (left), it becomes apparent that for very high tuna biomass ($y \geq 100 \si{\tonne}$) the model systematically underestimates tuna biomass. This result fits well with the improvement mentioned previously of the threshold regression task in relation to the standard regression. For this model, the MAE over set events drops down to $18.33$ tons, and over deployments it decreases also to $1.18$ tons. However, even with this threshold the model tends to underestimate tuna biomass when observed biomass is high (Figure~\ref{fig:reg_errors}, right). The marginal distributions for observed and estimated tuna biomass on set events are depicted in Figure~\ref{fig:density_dist}. Some possible factors that explain this underestimation are given in Section~\ref{sec:discussion}.


\section{Discussion}\label{sec:discussion}

The purpose of this paper is to introduce \textsc{Tun-AI}: a machine learning model for predicting tuna presence, and/or amount, under echo-sounder buoys attached to dFADs. To achieve this, we tested the performance of classification and regression methods, as well as the relative impact of including different levels of data on model performance. The approach used in the current study differs from previous work in that several models were tested for each task, and our results show that Gradient Boosting was the best model across all methods. This contrasts with the algorithms and methods applied by previous studies using echo-sounder data from dFAD buoys to predict the size of tuna aggregations ~\citep{Baidai2020MachineData, Orue2019FromBuoys, Lopez2016ADevices}. However, the assumptions and data-processing methods applied in other work may not be directly comparable to the process described here. For example,~\citet{Orue2019FromBuoys} or~\citet{Lopez2016ADevices} assume that tunas only occupy layers deeper than 25 m, thus omitting biomass estimates from shallower layers in their analysis. In our case, all layers were considered, as skipjack tuna are known to prefer warmer surface waters in areas where the thermocline is shallow~\citep{Andrade2003TheAtlantic}. In fact, later studies using the same approach as Lopez et al. (2016) did not achieve significant improvements on biomass estimates ~\citep{Orue2019AggregationSpecies}. When developing tuna presence/absence and classification models, ~\citet{Baidai2020MachineData} also chose to consider all layers in their analyses, which used data from a different brand of echo-sounder buoys in the Atlantic and Indian oceans, but did not consider oceanographic parameters in their models.

Our analysis also evaluated the impact of oceanographic conditions and position-derived variables on model performance. Across all tasks and models, the inclusion of additional features improved scores. This highlights the importance of enriching biomass estimates with contextual information when using data from echo-sounder buoys attached to dFADs. Previous studies have investigated the relationship between tropical tuna distribution and oceanographic conditions, both through catch data from observer logbooks and from dFAD data. For instance, in the Atlantic and Indian oceans skipjack tuna has been known to aggregate around upwelling systems and productive features where feeding habitat is favorable, and variables such as sea surface temperature or sea surface height have been shown to have a significant relation with tuna distribution ~\citep{Druon2017SkipjackOceans, Lopez2017EnvironmentalBuoys}. In addition, Spanish fishers using echo-sounder buoys on dFADs consider that the oceanographic context of the dFAD, and the characteristics of each ocean influence the accuracy of biomass estimates provided by buoys ~\citep{Lopez2014TheFisheries}.

It is worth noting that the oceanographic variables included in the current study were at surface level only (except for thermocline depth and SSHa). However, given the fact that tuna distribution within the water column is largely temperature dependent~\citep{Aoki2020PhysiologicalWaters,Hino2019ChangesSize} it is likely that models would further improve when considering variables depth-wise. Models could be further enriched when considering dFAD soak time, which has been relevant in previous research, or presence/absence of bycatch species and other species of tuna ~\citep{Orue2019AggregationSpecies, Lopez2017EnvironmentalBuoys, Forget2015BehaviourTelemetry}. In the current study, species composition of the catch data was not considered. As the echo-sounder buoys used in this study calculate biomass estimates based on the target strength of skipjack tuna, it is likely that the presence of other tuna species such as bigeye, which has a lower target strength~\citep{Boyra2018TargetFADs,Boyra2019InDevices}, would impact biomass estimates from the echo-sounder buoys, contributing to errors within the models used to estimate aggregation size. Most traditional echo-sounder buoys do not currently differentiate between species when giving biomass estimates, though recent buoy models, such as the ISD+ buoys included in the study, provide a daily estimate of species composition together with biomass estimates. Although previous studies have highlighted the importance of considering species composition when estimating biomass~\citep{Moreno2019TowardsDevices,Santiago2016TowardsTT-BAI}, the information from these buoys has not yet been applied, and should be considered in future studies. 

In the case of classification models, the confusion matrices in Figure~\ref{fig:cm} showed that most cases where the model misclassified the tuna aggregation size were when biomass estimates were medium ($10 \si{\tonne} \leq y < 30 \si{\tonne}$) or high ($y \geq 30 \si{\tonne}$). On the other hand, when examining the regression models we find that estimated tuna biomass tended to be lower than observed tuna biomass as the latter increased (Figure~\ref{fig:reg_errors}). This could be due to various factors: firstly, catches over 100 tons are relatively rare (in our data, \num{315} events, $8.1\%$) and thus the model does not have sufficient examples to properly learn from them; secondly, dFAD buoys are only able to estimate the biomass of tuna within the echo-sounder beam, and in tuna aggregations over 100 tons it is unlikely that the entire school is under the buoy at the same time. To resolve this issue, it could be interesting to apply specialist models which could be adjusted according to when aggregations are predicted to be small or large. 
Lastly, it is worth noting that fishermen do not choose on which buoys they set at random, but based on the raw biomass estimation provided to them, and thus could be biased towards buoys with higher biomass estimations. This could be a further reason why our ML models underestimate the observed tuna biomass when its values are above  $30 \si{\tonne}$. Future studies exploring the reasons behind fishermen’s decisions to visit a buoy could provide further insight into this point. This tendency to underestimate should also be taken into account when using information derived from echo-sounder buoys for stock assessments ~\citep{Santiago2016TowardsTT-BAI}, although consistent underestimation should have no effect on patterns present in the temporal series.

The potential applications of accurately quantifying tuna presence or absence around dFADs, as well as school aggregation size, are numerous. As highlighted by previous authors, data could be used for fishery-independent abundance indices, improving knowledge on species distribution or better understanding the factors driving aggregation and disaggregation processes of tuna at dFADs ~\citep{Santiago2016TowardsTT-BAI, Lopez2016ADevices,Moreno2019TowardsDevices}. The current study represents an important step in this direction, having successfully evaluated the performance of numerous models on achieving tasks of varying levels of complexity with high degrees of accuracy. As evidenced here and in previous research, biomass estimates from echo-sounder buoys in and of themselves provide approximate information on real tuna abundances around dFADs~\citep{Lopez2014EvolutionOceans}. However, when this massive data is further enriched with remote-sensing data on conditions throughout the water column and across oceans, and trained with reliable ground-truthed data, Machine Learning proves a powerful tool for extracting otherwise hidden patterns in the information.

\section*{Acknowledgments}

This study has been conducted using E.U. Copernicus Marine Service Information. We also thank AGAC for providing the logbook data used in the analysis and the helpful comments about the manuscript. The authors would also like to thank Carlos Roa for rendering available the Satlink echosounder dataset. The research of DGU has been supported in part by the Spanish MICINN under grants PGC2018-096504-B-C33 and RTI2018-100754-B-I00,  the European Union under the 2014-2020 ERDF Operational Programme and the Department of Economy, Knowledge, Business and University of the Regional Government of Andalusia (project FEDER-UCA18-108393). The research of Manuel Navarro-García has been financed by the research project IND2020/TIC-17526 (Comunidad de
Madrid).

\bibliographystyle{apalike}
\bibliography{references}

\newpage
\appendix
\appendixpage
\addappheadtotoc
\section{Classification task categories}

\begin{figure}[htbp]
    \centering
    \includegraphics[width=.8\linewidth]{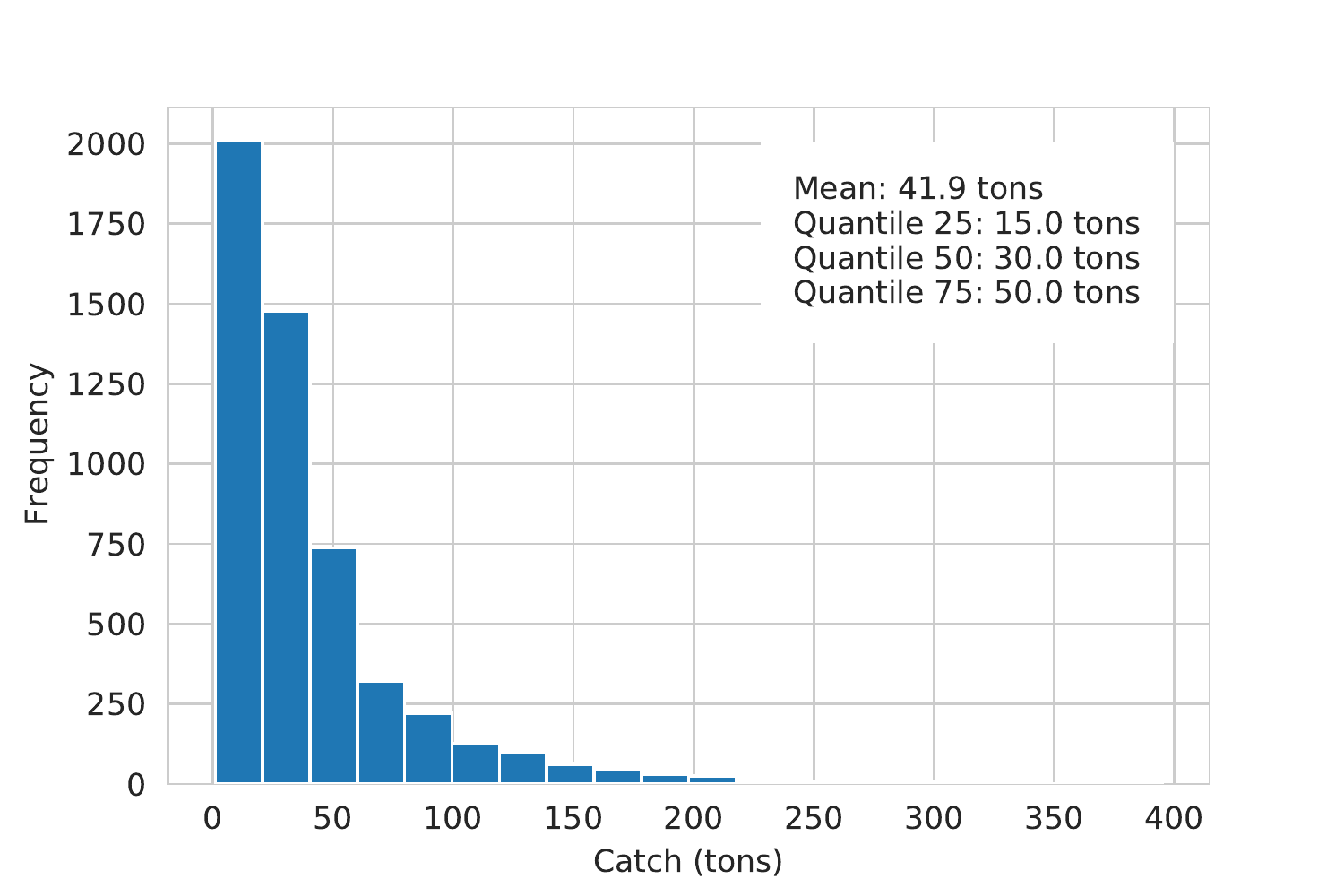}
    \caption{Distribution of caught tons of tuna, from a total of \num{5202} set events.}
    \label{fig:set_distribution}
\end{figure}

When defining the categories for the classification task, the thresholds were chosen according to various criteria. In both binary and ternary tasks, the lower threshold was based on best-practice guidelines to decrease shark bycatch, which recommend avoiding sets on tuna schools less than $10$ tons~\citep{Restrepo2016MitigationFisheries}. In the ternary classification task, the class limit for \textit{medium} was established according to the median catch ($30$ tons) in the dataset (see Figure~\ref{fig:set_distribution}).

\section{Feature importance}

An always challenging task in ML projects is their interpretability: to understand to some extent which features is the model mostly paying attention to. There exist several approaches to assess feature importance, and in this work we employ permutation importance \citep{Breiman2001RandomForests}, which evaluates the importance of a given feature as the drop in model's efficiency when the values of that column are shuffled randomly in the training set.

In Table~\ref{tab:feat_importance} we rank the 10 most important features for each task, based on the GB model. We explain next the meaning of each feature:
\begin{itemize}
    \item \texttt{Agg.T} is the aggregated value of the $72\times10$ echo-sound array.
    \item \texttt{N\_NaN} is the number of missing records, ranging from 0 to 72.
    \item \texttt{Agg.LY} is obtained by aggregating the 72 values at layer \texttt{Y}, where layer 1 is the closest to the surface.
    \item \texttt{Agg.HX} is obtained by aggregating the 10 values at hour \texttt{X}, where 0 is the closest to the event.
    \item \texttt{O2.X} and \texttt{Zos.X} are the dissolved oxygen and sea surface height, respectively, at hour \texttt{X}. As oceanography features are recorded daily, \texttt{X} can only take values  0, 23, 47 and 71 for these two variables.
    \item \texttt{Latitude} and \texttt{Longitude} are the buoy coordinates at hour 0.
    \item \texttt{Ocean} is a categorical variable that indicates the ocean where the event took place.
\end{itemize}

\begin{table}[htbp]
\centering
\small
\begin{tabular}{cllll}
\toprule
       & \multicolumn{2}{l}{Classification} & \multicolumn{2}{l}{Regression}\\
       \cmidrule(lr){2-3} \cmidrule(lr){4-5}
Rank & Binary & Three class & Standard & Threshold \\
\midrule
1 & \textcolor{ballblue}{\texttt{Agg.L5}}  & \textcolor{ballblue}{\texttt{Agg.L5}} & \textcolor{ballblue}{\texttt{Agg.T}}   & \textcolor{ballblue}{\texttt{Agg.T}} \\
2 & \textcolor{candypink}{\texttt{Longitude}}     & \textcolor{candypink}{\texttt{Longitude}}    & \textcolor{ballblue}{\texttt{Agg.L5}} & \textcolor{ballblue}{\texttt{Agg.L5}} \\
3 & \textcolor{candypink}{\texttt{Ocean}}         & \textcolor{candypink}{\texttt{Ocean}}        & \textcolor{candypink}{\texttt{Longitude}}    & \textcolor{candypink}{\texttt{Longitude}} \\
4 & \textcolor{ballblue}{\texttt{N\_NaN}}        & \textcolor{ballblue}{\texttt{Agg.L2}} & \textcolor{candypink}{\texttt{Latitude}}     & \textcolor{ballblue}{\texttt{Agg.L2}} \\
5 & \textcolor{candypink}{\texttt{Latitude}}      & \textcolor{ballblue}{\texttt{Agg.H10}} & \textcolor{ballblue}{\texttt{Agg.H32}} & \textcolor{sapgreen}{\texttt{Zos.0}} \\
6 & \textcolor{ballblue}{\texttt{Agg.L7}}  & \textcolor{sapgreen}{\texttt{Zos.0}} & \textcolor{ballblue}{\texttt{Agg.L2}} & \textcolor{candypink}{\texttt{Latitude}} \\
7 & \textcolor{ballblue}{\texttt{Agg.L1}}  & \textcolor{candypink}{\texttt{Latitude}}     & \textcolor{sapgreen}{\texttt{Zos.0}}        & \textcolor{ballblue}{\texttt{Agg.L6}} \\
8 & \textcolor{ballblue}{\texttt{Agg.L6}}  & \textcolor{sapgreen}{\texttt{Zos.71}}       & \textcolor{ballblue}{\texttt{Agg.H35}} & \textcolor{sapgreen}{\texttt{Zos.23}} \\
9 & \textcolor{ballblue}{\texttt{Agg.T}}    & \textcolor{sapgreen}{\texttt{O2.23}}        & \textcolor{sapgreen}{\texttt{Zos.23}}       & \textcolor{sapgreen}{\texttt{Zos.71}} \\
10 & \textcolor{ballblue}{\texttt{Agg.L2}} & \textcolor{sapgreen}{\texttt{O2.0}}         & \textcolor{sapgreen}{\texttt{Zos.71}}       & \textcolor{ballblue}{\texttt{Agg.H10}} \\
\bottomrule\\
\end{tabular}
\caption{Top ten most important features for the GB model in each of the 4 tasks. The variables are coloured depending on the feature group they belong. \textcolor{ballblue}{\texttt{Blue}} corresponds to acoustic features, \textcolor{sapgreen}{\texttt{green}} refers to oceanografic variables and \textcolor{candypink}{\texttt{red}} identifies geographical coordinates (or features derived from them)}
\label{tab:feat_importance}
\end{table}

Interpretation of feature importance must be exercised with care, since there are clearly correlations among the variables, and this has to be taken into account when interpreting permutation importance.

\section{Hyper-parameter search}

A grid search with cross-validation has been conducted to find the best hyper-parameters for each model by maximizing the ROC AUC. The hyper-parameter grid for all the models using the full set of features in each of the four tasks is shown in Tables \ref{tab:grid_RF}, \ref{tab:grid_XGB} and \ref{tab:grid_GB}. For Logistic Regression and Elastic Net, a grid search was conducted using the standard classes \texttt{LogisticRegressionCV} and \texttt{ElasticNetCV}, with a \texttt{l1\_ratio} grid of [0.0, 0.2, 0.4, 0.6, 0.8, 1.0] and [0.1, 0.5, 0.9, 1.0] respectively. The final values selected by cross-validation were 1 and 0.9. The name of the hyper-parameters coincide with the names that they have in Scikit-Learn's \citep{Pedregosa2011Scikit-learn:Python}the implementation. All the parameters not shown here were set to its default value (see the documentation for more details). Finally, the set of optimal hyper-parameters for each of the models is shown in Tables, \ref{tab:hyper_parametersRF}, \ref{tab:hyper_parametersXGB} and \ref{tab:hyper_parametersGB}.

\begin{table}[htbp]
\centering
\small
\begin{tabular}{lllll}
\toprule
Parameter  & Classification & Regression \\
\midrule
\texttt{n estimators}      &  [200, 500, 1000] & [100, 200, 500] \\
\texttt{max samples}     & [None, 0.8] & [None, 0.8]  \\
\texttt{max depth}         & [None, 2, 4] & [None, 4, 8]  \\
\texttt{min samples split} & [2, 8, 32] & [2, 8, 32]    \\
\texttt{min samples leaf}  & [1, 4, 16] & [1, 4, 16]    \\
\texttt{max features}      & [None, sqrt, log2] & [None, sqrt, log2]  \\
\bottomrule\\
\end{tabular}
\caption{Grid of hyper-parameters used in the Random Forest models.}
\label{tab:grid_RF}
\end{table}

\begin{table}[htbp]
\centering
\small
\begin{tabular}{lllll}
\toprule
Parameter  & Classification & Regression \\
\midrule
\texttt{n estimators}      &  [50, 100, 200] & [400] \\
\texttt{learning rate}     & [0.01, 0.1, 0.2] & [0.01, 0.1, 0.2]  \\
\texttt{max depth}         & [None, 3, 6] & [None, 3, 6] \\
\texttt{min samples split} & [2, 4, 8] & [2, 4, 8]   \\
\texttt{min samples leaf}  & [1, 2, 4] & [1, 2, 4]    \\
\texttt{max features}      & [None, sqrt, log2] & [None, sqrt, log2]  \\
\bottomrule\\
\end{tabular}
\caption{Grid of hyper-parameters used in the Gradient Boosting models.}
\label{tab:grid_GB}
\end{table}

\begin{table}[htbp]
\centering
\small
\begin{tabular}{lllll}
\toprule
Parameter  & Classification & Regression \\
\midrule
\texttt{n estimators}      & [50] & [50, 100, 200] \\
\texttt{learning rate}     & [0.2] & [0.01, 0.1, 0.2]  \\
\texttt{max depth}         & [2, 4] & [2, 4, 6]  \\
\texttt{subsample}         & [1.0] & [0.7, 1.0]    \\
\texttt{colsample bytree}  & [1.0] & [0.5, 1.0]    \\
\bottomrule\\
\end{tabular}
\caption{Grid of hyper-parameters used in the XGBoost models.}
\label{tab:grid_XGB}
\end{table}

\begin{table}[htbp]
\centering
\small
\begin{tabular}{lllll}
\toprule
       & \multicolumn{2}{l}{Classification} & \multicolumn{2}{l}{Regression}\\
       \cmidrule(lr){2-3} \cmidrule(lr){4-5}
Parameter & Binary & Three class & Standard & Threshold \\
\midrule
\texttt{n estimators}      &  1000 & 500 & 100 & 200  \\
\texttt{max samples}     & None & None &  None & None \\
\texttt{max depth}         & None & None & None & None \\
\texttt{min samples split} & 2 & 8 & 8   & 2    \\
\texttt{min samples leaf}  & 1 & 1 & 4  & 4   \\
\texttt{max features}      & sqrt & sqrt & None & None \\
\bottomrule\\
\end{tabular}
\caption{Best hyper-parameters for the Random Forest models trained on all features.}
\label{tab:hyper_parametersRF}
\end{table}

\begin{table}[htbp]
\centering
\small
\begin{tabular}{lllll}
\toprule
       & \multicolumn{2}{l}{Classification} & \multicolumn{2}{l}{Regression}\\
       \cmidrule(lr){2-3} \cmidrule(lr){4-5}
Parameter & Binary & Three class & Standard & Threshold \\
\midrule
\texttt{n estimators}      &  200 &  200 &  400 &  400 \\
\texttt{learning rate}     &  0.2 &  0.1 & 0.01 & 0.01 \\
\texttt{max depth}         &    6 &    6 & None & None \\
\texttt{min samples split} &    8 &    2 &    2 &    4 \\
\texttt{min samples leaf}  &    4 &    2 &    4 &    8 \\
\texttt{max features}      & log2 & log2 & auto & auto \\
\bottomrule\\
\end{tabular}
\caption{Best hyper-parameters for the Gradient Boosting models trained on all features.}
\label{tab:hyper_parametersGB}
\end{table}

\begin{table}[htbp]
\centering
\small
\begin{tabular}{lllll}
\toprule
       & \multicolumn{2}{l}{Classification} & \multicolumn{2}{l}{Regression}\\
       \cmidrule(lr){2-3} \cmidrule(lr){4-5}
Parameter & Binary & Three class & Standard & Threshold \\
\midrule
\texttt{n estimators}      & 50 & 50 & 200 & 200  \\
\texttt{learning rate}     & 0.2 & 0.2 & 0.01 & 0.01 \\
\texttt{max depth}         & 4 & 4 & 6 & 6 \\
\texttt{subsample}         & 1.0 & 1.0 & 0.7 & 0.7    \\
\texttt{colsample bytree}  & 1.0 & 1.0 & 0.5 & 0.5 \\
\bottomrule\\
\end{tabular}
\caption{Best hyper-parameters for the XGBoost models trained on all features.}
\label{tab:hyper_parametersXGB}
\end{table}

\end{document}